# An Efficient Technique for Similarity Identification between Ontologies

Amjad Farooq, Syed Ahsan and Abad Shah

**Abstract** -Ontologies usually suffer from the semantic heterogeneity when simultaneously used in information sharing, merging, integrating and querying processes. Therefore, the similarity identification between ontologies being used becomes a mandatory task for all these processes to handle the problem of semantic heterogeneity. In this paper, we propose an efficient technique for similarity measurement between two ontologies. The proposed technique identifies all candidate pairs of similar concepts without omitting any similar pair. The proposed technique can be used in different types of operations on ontologies such as merging, mapping and aligning. By analyzing its results a reasonable improvement in terms of completeness, correctness and overall quality of the results has been found.

————————— ◆ —————————

## 1. INTRODUCTION

It is quite difficult to retrieve relevant information from current web due to semantic heterogeneity problem in addition to other problems. Semantic web suggested solution of retrieval of specific information through ontologies, but ontologies may themselves suffer the heterogeneity problem when they are integrated, merged, shared, etc [1]. Same concept may be given different names or may be defined in different ways in two ontologies, although both ontologies belong to the same domain and may overlap. In order to use them together for different purposes such as merging, integrating, querying or even in creating a new ontology, those need to be aligned [2].

There are several techniques for aligning ontologies. They are mainly grouped into two classes: schema-based techniques and instance-based techniques. In schema-based techniques, the similarity among concepts of both ontologies is measured at structure-level while ignoring their actual data, whereas in instance-level techniques the similarity decision is made by taking actual data into consideration [3]. Ontology alignment at schema-level has different classifications such as structural, semantic, terminological and extensional [4]. Techniques for structural alignment are further divided into two classes: External structural alignment techniques and internal structural alignment techniques. External structure of a concept consists of the following elements [2], [5], [6]: super concepts; sub-concepts, sibling concepts and its non-taxonomic relations with the concepts. Therefore, when a concept $C_i$ of ontology A is aligned with a concept $C_j$ of ontology B, then all these elements i.e. super, sub and siblings are taken into consideration in the external-structure level alignment of a concept. Structure-level similarities between concepts of ontologies are computed using different criteria [7],[8], [9], [10] such as their direct super-entities (or all of their super-entities) are already similar; their sibling-entities (or all of their sibling-entities) are already similar; their direct sub-entities (or all of their sub-entities) are already similar; all (or most) of their descendant-entities (entities in the sub-tree rooted at the entity in question) are already similar; all (or most) of their leaf-entities (entities, which have no sub-entity, in the sub-tree rooted at the entity in question) are already similar and all (or most) of entities in the paths from the root to the entities in question are already similar.

While aligning different ontologies using criteria as mentioned above, we observe that certain pairs of similar concepts remain unaligned because those concepts don't satisfy these criteria .e. their respective surrounding concepts are not similar. Secondly, the matching process should be made at domain vocabulary declaration level, to make it more generic, simple and efficient. Since domain vocabulary is the foundation of ontology, therefore it is easy and efficient to make an alignment between domain vocabularies of two ontologies. In the alignment process, first domain vocabularies of both ontologies are enriched by equipping each domain concept with its possible synonyms and then the similarities between concepts of both vocabularies are determined and formalized. Finally, both ontologies may be adapted accordingly.

*Amjad Farooq, Syed Ahsan and Abad Shah: Computer Science and Engineering Department, University of Engineering and Technology, Lahore –Pakistan.*

Our proposed technique computes the similarity between concept $C_i$ of Ontology A and concept $C_j$ of ontology B, based on the criteria: (i) At least one of the super-concepts of $C_i$ and $C_j$ must be similar; (ii) Similarity between sub-concepts of $C_i$ and $C_j$ is optional; (iii) Similarity between siblings of $C_i$ and $C_j$ is optional; (iv) Similarity between non-taxonomic relations of both concepts in their ontologies is also optional.

However, to rank the level of similarity between those concepts or to measure their granularity level the options (ii), (iii) and (iv) may be considered.

Our proposed technique is more generic, simple and efficient because when two ontologies are aligned using this technique, none of similar pair of concepts remains unaligned. However existing techniques may leave some similar concepts unaligned. Moreover, the computing of optional similarities such as for sub-concepts, sibling concepts and non-taxonomic relations of both concepts involved in similarity computing process are omitted in this technique.

The remaining paper is organized as follows: In Section 2 an overview of some existing techniques is given. The proposed technique is given in Section 3 and it is illustrated by a case study in Section 4. The results generated through our technique and some existing techniques are compared in the same Section. Finally, the paper is concluded with future work in Section 5.

## 2. RELATED WORK

Semantic web heavily relies on ontologies. When ontologies need to be merged, integrated, queried and are used for knowledge sharing, their heterogeneity becomes bottleneck. To resolve this problem these need to be aligned [3, 9]. For alignment, the semantic similarities among the concepts of ontologies are determined. While determining similarities; the semantics, structural, taxonomical and contextual perspectives of concepts are taken into consideration [5]. The structure-level techniques for ontology alignment compare similarity of concepts based on different criteria as listed in previous Section.

According to [8], two concepts are similar if their direct super-entities are similar. In [11],[12],[13],[14] it is stated that two concepts are similar if their direct super-entities are similar, their sibling-entities are similar; their direct sub-entities are similar; their descendant-entities are similar; their leaf-entities are similar and entities in the paths from the root to those concepts are similar.

In [15], the structural similarity between two concepts is computed from the average similarity of their respective super and sub-concepts. The super and sub concepts of two concepts being compared are fetched into separate sets and then the resultant similarity is computed from the similarities of concepts in those sets. If both the super and sub concepts similarity are undefined, then the concepts being compared are declared as dissimilar otherwise they are declared similar concepts.

These techniques need to be more generic because for certain scenario these techniques are not suitable. For example, concepts of two ontologies are still similar (see Fig. 6 and Fig 8 in Sec. 4), although their super concepts, sibling-concepts, sub-concepts, descendent-concepts, leaf-concepts and concepts from root to those concepts are not similar. Furthermore, it is recommended that structural equality is not sufficient for measuring the alignment. Instead the concepts should be aligned on the basis of their semantics equality [15]. We think that these criteria should be more generic for computing similarities in two concepts.

## 3. PROPOSED TECHNIQUE

As mentioned earlier that our proposed technique computes the similarity between concepts of two ontologies based on criteria. At least one super-concept from both concepts involved in the matching process must be similar. There is no need of finding similarity between sub-concepts, siblings and their interactions with other concepts in their respective ontologies. Here we include only algorithm for extracting super-concepts and their matching. Algorithms for matching the siblings-concepts, sub-concepts and other interacting objects of both concepts involved in similarity measuring process are omitted here because according to our proposed criteria, all these are optional. However to rank the level of similarity, we are also working on these aspects.

Our proposed technique works in three steps: (i) Concepts Extraction (ii) Super-concepts Extraction (iii) Matching. This technique works with assumption that both ontologies are defined in Web Ontology Language (OWL) [16].

**Step 1: Concepts Extraction**

As stated earlier, only concepts are involved in alignment process based on external-structure whereas properties are involved in internal-structure alignment process. Therefore we extract concepts only and there is no need of extracting properties. Both source and target ontologies involved in alignment process are parsed and the concepts presented in "owl:CLASS" tag are extracted for determining their similarities. This step is very simple. Two vectors VA and VB are declared and then are populated from concepts of ontologies A and B respectively.

**Step 2: Super-Concepts Extraction**

Since the proposed criterion is based on the comparison of super-concepts for both concepts involved in the matching process, therefore for both ontologies we need all super-concepts of each concept extracted in previous step. Working of this step in pseudo form is given in Fig 1. Vectors named as CSA (Concepts along with Super-concepts of ontology

```
Algorithm: Super-concept-Extraction
Input: Ontology A, Ontology B
Output: Vector CSA, Vector CSB
BEGIN
(i)   Declare a vector two CSA and CSB to store concepts along with
      their super-concepts of A  and B respectively; where CSA and CSB
      are of  structure type with two elements, one for concept and
      other for its respective super-concept.
(ii)  Populate a vector CSA from each concept and its immediate
      super-concept of A.
(iii) FOR  each cA in VA
              superVector = CSA.getSupers(cA)
                  CSA.add(cA , superVector)
          NEXT
(iv)  Populate a vector CSB from each concept and its immediate super-
      concept of B.
(v)    FOR each cB in VB
              superVector = CSB.getSupers(cB)
                  CSB.add(cB, superVector)
        NEXT
   END
```

Fig 1. Super-concepts extraction from Ontologies A and B

A) and CSB(Concepts along with Super-concepts of ontology B) are obtained as output of this step.

**Step 3: Matching**

We claim that ontology alignment based on similarities of external structure of concepts need only the similarity of at least on super-concept (immediate is not necessarily) from both concepts involved in matching process. Therefore in this step, each super-concept of cA presented in VA is compared with each super-concept of $cB_i$ for all $1 \leq i \geq n$. If there is a match between any super-concept of cA and any super-concept of cB then both cA and cB are declared similar concepts. A working of matching process in pseudo form is repented in Fig 2. The output e.g. SimilarPair Vector will consist of only those pairs of concepts having similarities from both ontolgies.

## 4. CASE STUDY

The evaluation of proposed technique was performed automatically and manually with three test cases: (a) Both the source ontology A and the target ontology B belongs to same domain but are developed separately with different viewpoints. (b) Both the source ontology A and the target ontology B belongs to different domains with no similarity. (c)Using same ontology as the source ontology A and the target ontology B.

The algorithms listed in the previous section were implemented in Java programming language. A pair of ontologies *developed by different persons* were used for evaluation. In first pair (see Fig.3), both ontologies were about "Research Activities" domain conducted in a university. Workings of proposed technique were also traced manually, using same pairs of ontologies. Then the results obtained through automatically and manually were compared with each other and with expected results mentioned by respective domains experts. We found that results were absolutely correct and complete.

```
Algorithm: SimilarityFinding
Input: Vector CSA, Vector CSB
Output: Vector SimilarPair
BEGIN
(i) Declare a vector SimilarPair to store similar pairs of concepts, where similarPair
        is a structure with two elements for concept cA and other of concept cB.
(ii) FOR each cA in CSA
                Same =false
                FOR each cB in CSB
                Same=isAnySuperSame(cA.superVector,cB.superVector)
                        IF same THEN
                                SimilarPair.add(cA,cB)
                        ENDIF
                NEXT
        NEXT
        FUNCTION isAnySuperSame(Vector V1 , Vector V2):Boolean
        match=false
        FOR each item1 in V1
                FOR each item2 in V2
                        IF item1=item2 THEN
                                match=true
                        END IF
                NEXT
        NEXT
        RETURN match
        END function
END
```

Fig 2. Concepts matching process

```xml
<owl:Class rdf:ID="Article">
  <rdfs:subClassOf rdf:resource="#Publication"/>
</owl:Class>
<owl:Class rdf:ID="AssistantProfessor">
  <rdfs:subClassOf rdf:resource="#Faculty"/>
</owl:Class>
<owl:Class rdf:ID="AssociateProfessor">
  <rdfs:subClassOf rdf:resource="#Faculty"/>
</owl:Class>
<owl:ObjectProperty rdf:ID="authorOf">
  <rdfs:domain rdf:resource="#Faculty"/>
  <owl:inverseOf rdf:resource="#hasAuthor"/>
</owl:ObjectProperty>
<owl:ObjectProperty rdf:ID="belongsTo">
  <rdfs:domain rdf:resource="#Faculty"/>
  <rdfs:range rdf:resource="#EducationalOrganization"/>
  <owl:inverseOf rdf:resource="#hasFaculty"/>
</owl:ObjectProperty>
<owl:Class rdf:ID="BookChapter">
  <rdfs:subClassOf rdf:resource="#Article"/>
</owl:Class>
<owl:Class rdf:ID="College">
  <rdfs:subClassOf rdf:resource="#EducationalOrganization"/>
</owl:Class>
<owl:Class rdf:ID="ConferencePaper">
  <rdfs:subClassOf rdf:resource="#Article"/>
</owl:Class>
<owl:Class rdf:ID="Department">
  <rdfs:subClassOf rdf:resource="#University"/>
</owl:Class>
<owl:Class rdf:ID="EducationalOrganization">
```

```xml
<owl:Class rdf:ID="AffiliatedInstitute">
  <rdfs:subClassOf rdf:resource="#University"/>
</owl:Class>
<owl:Class rdf:ID="AssistantProfessor">
  <rdfs:subClassOf rdf:resource="#Professor"/>
</owl:Class>
<owl:Class rdf:ID="AssociateProfessor">
  <rdfs:subClassOf rdf:resource="#Professor"/>
</owl:Class>
<owl:ObjectProperty rdf:ID="belongsTo">
  <rdfs:domain rdf:resource="#Faculty"/>
  <rdfs:range>
    <owl:Class>
      <owl:unionOf rdf:parseType="Collection">
        <owl:Class rdf:about="#College"/>
        <owl:Class rdf:about="#School"/>
        <owl:Class rdf:about="#University"/>
      </owl:unionOf>
    </owl:Class>
  </rdfs:range>
  <owl:inverseOf rdf:resource="#hasFaculty"/>
</owl:ObjectProperty>
<owl:Class rdf:ID="BookChapter">
  <rdfs:subClassOf rdf:resource="#Publication"/>
</owl:Class>
<owl:Class rdf:ID="College">
  <rdfs:subClassOf rdf:resource="#Organization"/>
</owl:Class>
<owl:Class rdf:ID="Conference">
  <rdfs:subClassOf rdf:resource="#Publication"/>
</owl:Class>
<owl:Class rdf:ID="Department">
  <rdfs:subClassOf rdf:resource="#University"/>
</owl:Class>
<owl:Class rdf:ID="Employee">
```

Fig 3. Sample code slice of ontologies A and B

The hierarchical structure of concepts of both ontolgies is given in Fig 4.

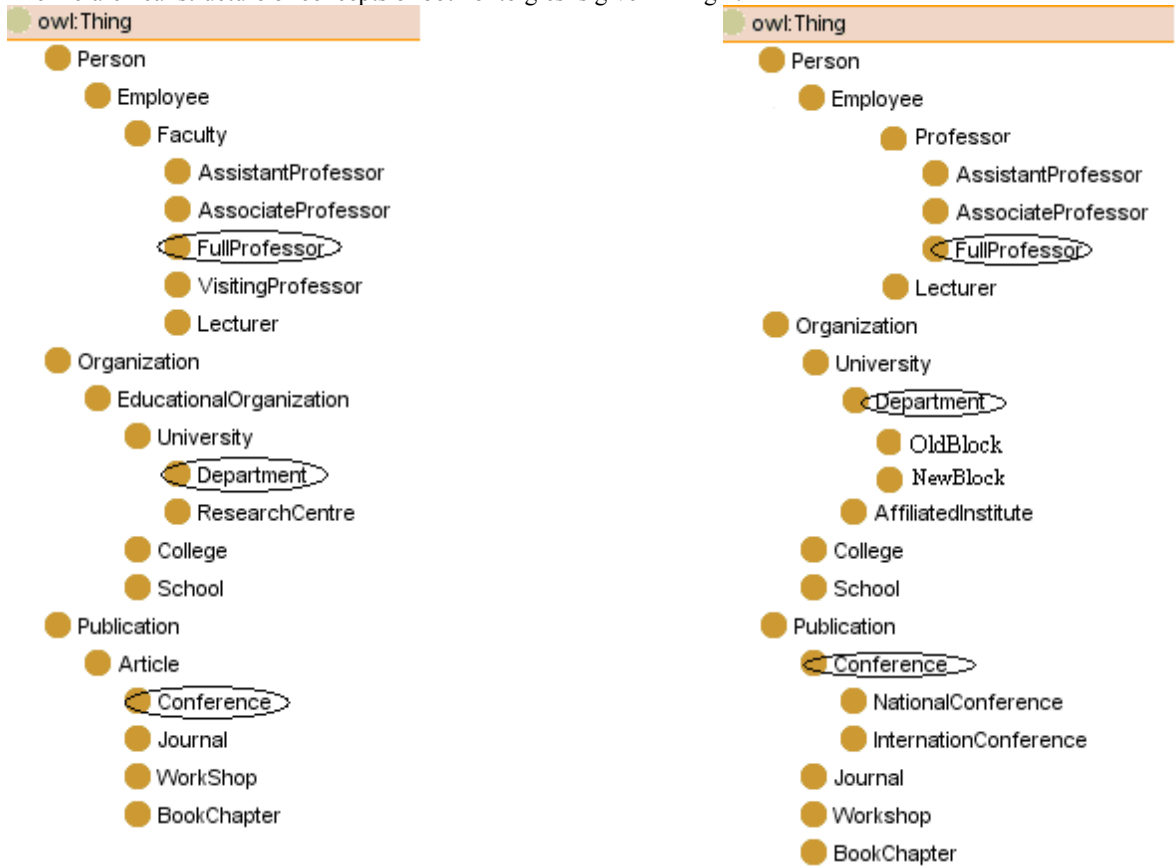

Fig 4. Slice of Hierarchical Presentation of Concepts of Ontologies A and B.

**Step 1:**

For manual verification of we took a few concepts from both ontologies. The concepts encircled in Figure 6 were chosen for determining their similarities for alignment process

**Step 2:**

Assume that concept ID of FullProfessor concept of A is $c1\_A$.
SUPC of $c1\_A$ = [Faculty, Employee, Person]
Assume that concept ID of Department concept of A is $c2\_A$.
SUPC of $c2\_A$ = [Organization, EducationOrganization, University]
Assume that concept ID of Conference concept of A is $c3\_A$.
SUPC of $c3\_A$ = [Publication, Article]
Assume that concept ID of FullProfessor concept of B is $c1\_B$.
SUPC of $c1\_B$ = [Professor, Employee, Person]
Assume that concept ID of Department concept of B is $c2\_B$.
SUPC of $c2\_B$ = [Organization, University]
Assume that concept ID of Conference concept of B is $c3\_B$.

SUPC of $c3\_B$ = [Publication]

**Step 3:**

We organized the super-concepts, sub-concepts and siblings concepts of three sample concepts in a tabular form as given in Table 1.

● According to [8], (c1_A, c1_B) and (c3_A, c3_B) is not similar because their direct super-entities are not similar.

● According to [11],[12],[13],[17], (c2_A, c2_B) is not similar because their sub-entities are not similar and their sibling-entities are not similar.

● According to [11,12,13], (c2_A, c2_B) and (c3_A, c3_B) are not similar because their leaf-entities are not similar and entities in the paths from the root to these concepts are not similar. Whereas according to our proposed criterion (c1_A, c1_B),(c1_A, c1_B) and (c1_A, c1_B) are similar and allthese concepts are actually similar, determined through manual matching and with results provided by respective domain experts.

Tab 1. concepts with their surroundings concepts of A and B

| Concept_id | Concept | SUPC | SUBC | SBLC |
|---|---|---|---|---|
| C1_A | FullProfessor | Employee<br>Faculty<br>Person | --- | AssistantProfessor<br>AssociateProfessor<br>VisitingProfesssor<br>Lecturer |
| C1_B | FullProfessor | Professor<br>Employee<br>Person | ------- | AssistantProfessor<br>AssociateProfessor |
|  |  |  |  |  |
| C2_A | Department | Organization<br>EducationOrganization<br>University | ---------- | ResearchCentre |
| C2_B | Department | Organization<br>University | OldBlock<br>NewBlock | AffiliatedInstitute |
|  |  |  |  |  |
| C3_A | Conference | Publication<br>Article |  | JournalArticle<br>WorkshopPaper<br>BookChapter |
| C3_B | Conference | Publication | NationalConferencePaper<br>InternationalConferencePaper | JournalArticle<br>WorkshopPaper<br>BookChapter |

```
<owl:Class rdf:ID="BAStudent">
    <rdfs:subClassOf rdf:resource="#UnderGradStdudent"/>
  </owl:Class>
  <owl:Class rdf:ID="BComStudent">
    <rdfs:subClassOf rdf:resource="#UnderGradStdudent"/>
  </owl:Class>
  <owl:Class rdf:ID="BScStudent">
    <rdfs:subClassOf rdf:resource="#UnderGradStdudent"/>
  </owl:Class>
  <owl:Class rdf:ID="GradSudent">
    <rdfs:subClassOf rdf:resource="#Student"/>
  </owl:Class>
  <owl:Class rdf:ID="MAStudent">
    <rdfs:subClassOf rdf:resource="#GradSudent"/>
  </owl:Class>
  <owl:Class rdf:ID="MBAStudent">
    <rdfs:subClassOf rdf:resource="#GradSudent"/>
  </owl:Class>
  <owl:Class rdf:ID="MPhilStudent">
    <rdfs:subClassOf rdf:resource="#PostGradStudent"/>
  </owl:Class>
  <owl:Class rdf:ID="MScStudent">
```

```
<owl:Class rdf:ID="BAStudent">
    <rdfs:subClassOf rdf:resource="#UnderGradStdudent"/>
  </owl:Class>
  <owl:Class rdf:ID="BComStudent">
    <rdfs:subClassOf rdf:resource="#UnderGradStdudent"/>
  </owl:Class>
  <owl:Class rdf:ID="BScStudent">
    <rdfs:subClassOf rdf:resource="#UnderGradStdudent"/>
  </owl:Class>
  <owl:Class rdf:ID="GradSudent">
    <rdfs:subClassOf rdf:resource="#Student"/>
  </owl:Class>
  <owl:Class rdf:ID="HEC_Student">
    <rdfs:subClassOf rdf:resource="#PhDStudent"/>
  </owl:Class>
  <owl:Class rdf:ID="LocalStudent">
    <rdfs:subClassOf rdf:resource="#PhDStudent"/>
  </owl:Class>
  <owl:Class rdf:ID="MAStudent">
    <rdfs:subClassOf rdf:resource="#GradSudent"/>
  </owl:Class>
  <owl:Class rdf:ID="MBAStudent">
    <rdfs:subClassOf rdf:resource="#GradSudent"/>
```

Fig 7. Sample code slice of ontologies C and D generated in OWL

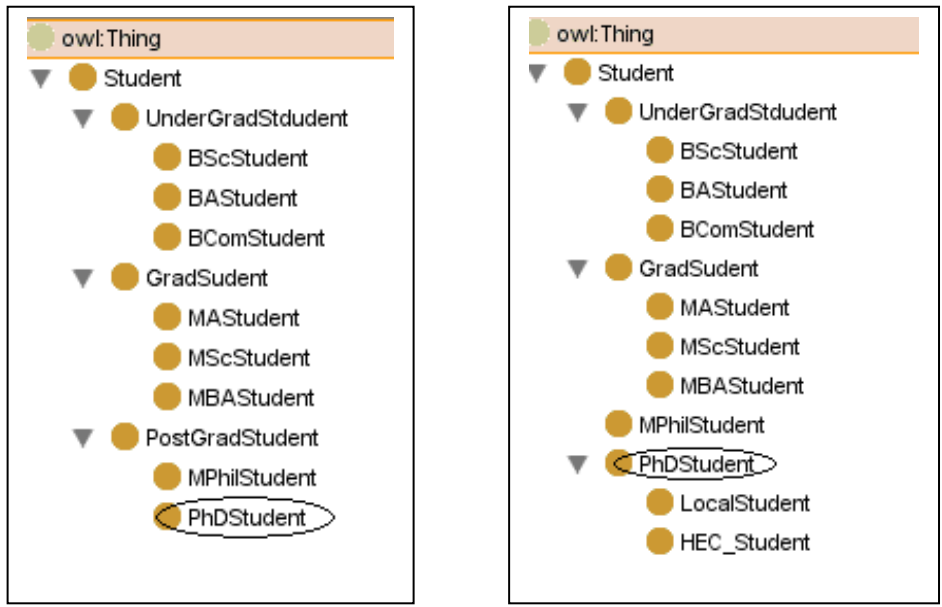

Fig 8: Slice of hierarchical presentation of concepts of Ontologies C and D.

To further verify proposed technique we have taken two independent ontologies C and D of Student domain, created by different groups. A sample code slice of both ontologies are shown in Figure 7.

**Step 1:**

For manual verification of we took a few concepts from both ontologies. The concepts encircled in Figure 8 were chosen for determining their similarities for alignment process

**Step 2:**

Assume that concept ID of PhDStudent concept of C is $C_i\_C$.
SUPC of $C_i\_C$ = [PostGradStudent, Student]
Assume that concept ID of PhDStudent concept of D is $C_j\_D$.
SUPC of $C_j\_D$ = [Student]

**Step 3:**

We organized the super-concepts, sub-concepts and siblings concepts of a sample concept in a tabular form as given in Table 2.

Tab 2. Concepts with their surroundings concepts of ontologies C and D

| Concept _id | Concept | SUPC | SUBC | SBLC |
|---|---|---|---|---|
| Ci_C | PhDStudent | PostGradStudent, Student | --- | MPhilStudent |
| Cj_D | PhDStudent | Student | LocalStudent, HEC_Student | MPhilStudent, GradStudent, UnderGradStudent |

Since Condition (Sec.3 (step 3, ii)) is true for concepts Ci_C and Cj_D, therefore the concepts ci_C and Cj_D are similar.
*Results with respect to existing approaches:*

- According to [8], (Ci_C, Cj_D) is *not similar* because their direct super-entities are not similar.

- According to [11],[17],[18], (Ci_C, Cj_D) is *not similar* because their sub-entities are not similar and their sibling-entities are not similar.

- According to [11],[20],[21], (Ci_C, Cj_D) is *not similar* because their leaf-entities are not similar and entities in the paths from the root to these concepts are not similar.
Whereas according to our proposed technique (Ci_C, Cj_D) is *similar* and these concepts are actually similar, determined through manual matching and from respective domain experts.

The proposed technique is tested with two ontologies from different domains with no similarity
The ontology from tourism domain (see Fig. 9) has taken as source ontology and ontology B of Research Activities

Tab 3: Results comparison with existing techniques

| Concepts(Ci, ,Cj) | Information Content[9] | OWL Lite Aligner [13] | *Anchor Prompt*[11] | Similarity Flooding [12] | Proposed Technique |
|---|---|---|---|---|---|
| (C1_A, C1_B) | X | X | X | X | √ |
| (C2_A, C2_B) | √ | √ | √ | √ | √ |
| (C3_A, C3_B) | X | √ | X | √ | √ |
| (Ci_C, Cj_D) | X | X | X | X | √ |

domain used in previous case has taken as target ontology. The correct result was found.

Proposed technique is verified using same ontology as a source and target. It was found that both ontologies i.e. source ontology A and target ontology B was absolutely aligned.

```
<owl:Class rdf:ID="Sunbathing">
<rdfs:subClassOf rdf:resource= "#Relaxation" /> </owl:Class>
<owl:Class rdf:ID="Sports">
<owl:disjointWith> <owl:Class rdf:about="#Adventure" />
</owl:disjointWith>
<owl:disjointWith> <owl:Class rdf:about="#Relaxation" />
l:disjointWith>
:disjointWith> <owl:Class rdf:about="#Sightseeing" />
l:disjointWith>
:subClassOf> <owl:Class rdf:ID="Activity" />
s:subClassOf> </owl:Class>
:Class rdf:ID="Yoga">
> <owl:Class rdf:about="#Relaxation" />
s:subClassOf> </owl:Class>
:Class rdf:ID="BudgetAccommodation">
:equivalentClass>
:Class> <owl:Class rdf:ID="LuxuryHotel">
:subClassOf> <owl:Class rdf:about="#Hotel" />
s:subClassOf>
:subClassOf> <owl:Class rdf:ID="Beach">
subClassOf rdf:resource="#Destination" />
l:Class> <owl:Class rdf:ID="Hotel">
:disjointWith> <owl:Class rdf:about="#BedAndBreakfast" />
l:disjointWith>
:disjointWith> <owl:Class rdf:about="#Campground" />
l:disjointWith>
:subClassOf rdf:resource="#Accommodation" />
l:Class> <owl:Class rdf:ID="Museums">
:subClassOf> <owl:Class rdf:ID="City"> <rdfs:subClassOf>
:Class rdf:about="#UrbanArea" />
s:subClassOf> </owl:Class>
:Class rdf:ID="Safari">
:subClassOf> <owl:Class rdf:about="#Adventure" />
```

Fig 9 Sample code slice of travel ontology

Our proposed technique is based on such criterion through which all similar concepts are determined and aligned. Results are concluded in previous section are summarized in Table 3. In table 3, the first column contains concepts Ids as declared in previous section. The cell-value (X) indicates that both concepts are not similar according to respective technique and the value (√) indicates that both concepts are similar.

## 5. CONCLUSION AND FUTURE WORK

In this paper we have proposed a very simple and generic technique to determine similarity between concepts of two ontologies. All super-concepts of each concept to be aligned and its interaction with other concepts of two ontologies were grouped into separate vectors. Then matching process was performed according to proposed algorithms and it was determined manually and through logic of algorithm that one concept of ontology A was similar with a concept of ontology B although there were some disparities in their respective super-concepts, sibling-concepts, sub-concepts and their interaction with other objects. The proposed technique can be extended to handle the granularity and the level of similarities. Primitive characteristics of concepts may also be incorporated while measuring their similarities.

## ACKNOWLEDGEMENT

This research work has been supported by the "Higher Education Commission of Pakistan", and the University of Engineering and Technology, Lahore.

## REFERENCES

[1]  J.Euzenat and P. Shvaiko, "Ontology matching, " Springer-Verlag, Heidelberg (DE), 333p., 2007.

[2]  P. Shvaiko, J. Euzenat, "A Survey of Schema-based Matching Approaches," Journal on Data Semantics, 2005.

[3]  P. Shvaiko, F. Giunchiglia, P. Silva, D. McGuinness, "Web Explanations for Semantic Heterogeneity Discovery," ESWC 2005: 303-317.

[4]  J. Euzenat, T. Bach, J. Barrasa, P. Bouquet, J. Bo, R. Dieng-Kuntz, M. Ehrig, M. Hauswirth, M. Jarrar, R. Lara, D. Maynard, A. Napoli, G. Stamou, H. Stuckenschmidt, P. Shvaiko, S. Tessaris, S. Acker, and I. Zaihrayeu, "State of the art on ontology alignment," Technical Report 2.2.3, Knowledge Web NoE, 2004.

[5]  M. Ehrig, A. Koschmider and A. Oberweis, "Measuring Similarity between Semantic Business Process Models," In John F. Roddick and Annika Hinze, Conceptual Modelling 2007, Proceedings of the Fourth Asia-Pacific Conference on Conceptual Modelling (APCCM 2007), Australia.

[6]  P. Lambrix, H. Tan, "SAMBO - A System for Aligning and Merging Biomedical Ontologies," Journal of Web Semantics, special issue on Semantic Web for the Life Sciences, 2006.

[7] J. Madhavan, P. Bernstein, and E. Rahm, "schema matching using cupid," Proc. of the 27th VLDB, 2001, pp. 48–58.

[8] P. Valtchev and J. Euzenat, "Dissimilarity measure for collections of objects and values," Lecture Notes in Computer Science. London, UK: Springer, 1997, vol. 1280.

[9] B. Hariri, H. Abolhassani and A. Khodaei, "A new Structural Similarity Measure for Ontology Alignment," Proceedings of the 2006 International Conference on Semantic Web & Web Services, SWWS 2006, Las Vegas, Nevada, USA.

[10] B. Chen, H. Tan and P. Lambrix, "Structure-Based Filtering for Ontology Alignment," 15th IEEE International Workshops on Enabling Technologies: Infrastructure for Collaborative Enterprises (WETICE'06), pp. 364-369, 2006.

[11] N. Noy and M. Musen, "Anchor-prompt: using non-local context for semantic matching," Proc. of the workshop on Ontologies and Information Sharing at the International Joint Conference on Artificial Intelligence (IJCAI), 2001, pp. 63–70.

[12] S. Melnik, H. Garcia-Molina, and E. Rahm, "A versatile graph matching algorithm," Proc. of ICDE, 2002.

[13] J. Euzenat and P. Valtchev, "An integrative proximity measure for ontology alignment," in Proceedings of Semantic Integration workshop at ISWC, 2003.

[14] M. Ehrig, P. Haase, N. Stojanovic and M. Hefke, "Similarity for Ontologies - A Comprehensive Framework," Proc. of *13th European Conference on Information Systems*. Regensburg, May 2005.

[15] J. Euzenat, "Semantic precision and recall for ontology alignment evaluation," Proc. 20th International Joint Conference on Artificial Intelligence (IJCAI), Hyderabad (IN), pp248-253, 2007

[16] Peter F. Patel-Schneider, Patrick Hayes, and Ian Horrocks eds. OWL Web Ontology Language Semantics and Abstract Syntax. W3C Recommendation, 10 February 2004.

[17] J. Zhong, H. Zhu, Y. Li, and Y. Yu, "Conceptual graph matching for semantic search," Proc. of Conceptual Structures: Integration and Interfaces (ICCS-2002), 2002, pp. 92–106.

[18] R. Yves J. Mary, E. Shironoshita, R. Mansur, "Ontology matching with semantic verification. Web Semantics," Science, Services and Agents on the World Wide Web ,2009, Volume 7 Issue 3, 235-251.

[19] R. Dieng and S. Hug, "Comparison of personal ontologies represented through conceptual graphs," *Proc. of the 13th ECAI*, 1998.

[20] J. Euzenat and P. Valtchev, "Similarity-based ontology alignment in OWL-lite," In *Proc. 15th ECAI*, pages 333–337, Valencia (ES), 2004.

[21] M. Ehrig and Y. Sure, "Ontology mapping - an integrated approach," Proc. of the European Semantic Web Symposium (ESWS), pages 76–91, 2004.